\pdfoutput=1
\documentclass[11pt]{article}
\usepackage[]{ACL2023}
\usepackage{times}
\usepackage{latexsym}
\usepackage[T1]{fontenc}
\usepackage[utf8]{inputenc}
\usepackage{multirow}
\usepackage{graphicx}
\usepackage{pgfplots}
\usepackage{pgf}
\usepackage{amsmath}
\usepackage{amssymb,mathrsfs}
\usepackage{array}
\usepackage[caption=false]{subfig}
\usepackage{tikz}
\usepackage{float}
\usepackage{makecell}
\usepackage{xcolor}
\usepackage{tcolorbox}
\usepackage{bm}
\usepackage{microtype}
\usepackage{xspace}
\usepackage{inconsolata}
\usetikzlibrary{chains,scopes,positioning,backgrounds,shapes,fit,shadows,calc,arrows.meta,matrix}

\title{TranSFormer: Slow-Fast Transformer for Machine Translation}

\author{
  Bei Li$^1$\thanks{\xspace\xspace The work was done when the first author was an intern at Microsoft Research Asia.},
  Yi Jing$^1$,
  Xu Tan$^2$,
  Zhen Xing$^3$,
  \textbf{Tong Xiao$^{1,4}$\thanks{\xspace\xspace Corresponding author.}} and \textbf{Jingbo Zhu$^{1,4}$}\\
  $^{1}$School of Computer Science and Engineering, Northeastern University, Shenyang, China\\
  $^{2}$Microsoft Research Asia, $^{3}$Fudan University\\
  $^{4}$NiuTrans Research, Shenyang, China\\
  {\tt
        \{libei\_neu,jingyi\_neu\}@outlook.com, xuta@microsoft.com
  }\\
  {\tt
        zxing20@fudan.edu.cn, \{xiaotong,zhujingbo\}@mail.neu.edu.cn
  }
}

\begin{document}
\maketitle
\begin{abstract}
Learning multiscale Transformer models has been evidenced as a viable approach to  augmenting machine translation systems. Prior research has primarily focused on treating subwords as basic units in developing such systems. However, the incorporation of fine-grained character-level features into multiscale Transformer has not yet been explored. In this work, we present a \textbf{S}low-\textbf{F}ast two-stream learning model, referred to as Tran\textbf{SF}ormer, which utilizes a ``slow'' branch to deal with subword sequences and a ``fast'' branch to deal with longer character sequences. This model is efficient since the fast branch is very lightweight by reducing the model width, and yet provides useful fine-grained features for the slow branch. Our TranSFormer shows consistent BLEU improvements (larger than 1 BLEU point) on several machine translation benchmarks.

\end{abstract}

\section{Introduction}
Transformer~\citep{vaswani2017attention} has demonstrated strong performance across a range of natural language processing (NLP) tasks. Recently, learning multiscale Transformer models has been evidenced as a promising approach to  improving standard Transformer. Previous research on this line can be broadly categorized into two streams: one learns local fine-grained features by using a fixed-length window ~\citep{yang-etal-2019-convolutional,hao-etal-2019-multi,guo-2020-msan}, linguistic-inspired local patterns \cite{li2022multiscale}, and a hybrid approach that combines convolution and self-attention models ~\citep{gulati2020conformer} or run in parallel~\citep{Zhao-2019-MUSE}; the other learns sequence representations by considering multiple subword segmentation/merging schemas \cite{wu2020mixed}.
\definecolor{tiffanyblue}{RGB}{129,216,208}
\definecolor{ured}{RGB}{255,160,112}
\definecolor{darkgreen}{RGB}{157, 184,2}
\definecolor{uyellow}{RGB}{239,202,102}
\definecolor{ublue}{RGB}{147, 176, 208}
\definecolor{violet}{RGB}{238, 130,238}
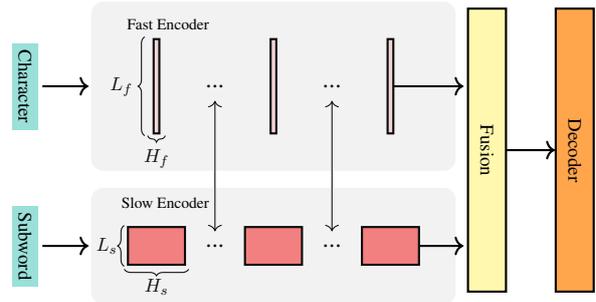
\begin{figure}[t!]
  \centering
  \begin{tikzpicture}
    \tikzstyle{every node}=[scale=0.65]
    \newlength{\seg}
    \setlength{\seg}{1.5em}     
    \newlength{\subseg}
    \setlength{\subseg}{50em}    
    \newlength{\resseg}       
    \setlength{\resseg}{1.5em}  
    \newlength{\shift}
    \setlength{\shift}{15em}     
    \newlength{\shiftMULLEFT}
    \setlength{\shiftMULLEFT}{15em}     
    \newlength{\shiftMULRIGHT}
    \setlength{\shiftMULRIGHT}{5em}     
    
    \newlength{\shiftSOFTLEFT}
    \setlength{\shiftSOFTLEFT}{15em}     
    \newlength{\shiftSOFTRIGHT}
    \setlength{\shiftSOFTRIGHT}{5em}     
    
    \newlength{\shiftMAPPINGLEFT}
    \setlength{\shiftMAPPINGLEFT}{15em}     
    
    \newlength{\shiftADDLEFT}
    \setlength{\shiftADDLEFT}{5em}     

    \newlength{\seggsan}
    \setlength{\seggsan}{1.7em}     
    \newlength{\subseggsan}
    \setlength{\subseggsan}{30em}    
    \newlength{\resseggsan}       
    \setlength{\resseggsan}{1.5em}  

    \tikzstyle{circlenode}=[circle,minimum size=4pt,inner sep=0];
    \tikzstyle{add}=[circle,minimum size=1.2em,inner sep=0];
    \tikzstyle{odenode1}=[draw,minimum height=5em,minimum width=0.3em,inner sep=0,thick];
    \tikzstyle{odenode2}=[draw,minimum height=2em,minimum width=3em,inner sep=0,thick];
    \tikzstyle{charinput}=[draw,minimum height=6em,minimum width=1em,inner sep=0,thick];
    \tikzstyle{subwordinput}=[draw,minimum height=4em,minimum width=1em,inner sep=0,thick];
    \tikzstyle{decodernode}=[draw,minimum height=15em,minimum width=2em,inner sep=0,thick];
    \tikzstyle{moduleode}=[draw,minimum height=1.2em,minimum width=3.5em,inner sep=0,thick,rounded corners=3.5pt];
    \tikzstyle{linecorner}=[rounded corners=3pt];

    
    

    \node [anchor=center,font=\footnotesize] (char) at (0em,0em){};
    \node [anchor=center,font=\footnotesize] (subword) at (0em,-5.5em){};

    \node[fill=tiffanyblue!80] (cinput) at ([xshift=-2em]char) {\rotatebox{270}{Character}};
    \node[fill=tiffanyblue!80] (sinput) at ([xshift=-2em]subword) {\rotatebox{270}{Subword}};



    \node[odenode1, fill=pink!50,anchor=center](lc1) at ([xshift=2.5em]char.center){};
    \node[odenode2, fill=red!50,anchor=center](ls1) at ([xshift=2.5em]subword.center){};

    \draw [thick,->] ([xshift=0.15em]cinput.east) -- ([xshift=-2.2em]lc1.west) ;
    \draw [thick,->] ([xshift=0.15em]sinput.east) -- ([xshift=-2.2em,yshift=-5.5em]lc1.west) ;

    \node[odenode1, fill=pink!50,anchor=center](lc2) at ([xshift=4em]lc1.center){};
    \node[odenode2, fill=red!50,anchor=center](ls2) at ([xshift=4em]ls1.center){};


    \node[odenode1, fill=pink!50,anchor=center](lc3) at ([xshift=4em]lc2.center){};
    \node[odenode2, fill=red!50,anchor=center](ls3) at ([xshift=4em]ls2.center){};

    \node[](dot1) at ([xshift=2em]lc1.center){\Large ...};
    \node[](dot2) at ([xshift=2em]ls1.center){\Large ...};

    \node[](dot1) at ([xshift=2em]lc2.center){\Large ...};
    \node[](dot2) at ([xshift=2em]ls2.center){\Large ...};
    
    \draw [<->] ([xshift=2em,yshift=-0.5em]lc1.center) -- ([xshift=2em,yshift=0.5em]ls1.center) ;

    \draw [<->] ([xshift=2em,yshift=-0.5em]lc2.center) -- ([xshift=2em,yshift=0.5em]ls2.center) ;

    \node[decodernode, fill=yellow!40,anchor=center](fusion) at ([xshift=3.3em,yshift=3.3em]ls3.center){\rotatebox{270}{Fusion}};

    \node[decodernode, fill=orange!70,anchor=center](decoder) at ([xshift=3em]fusion.center){\rotatebox{270}{Decoder}};

    \draw [thick,->] (lc3.east) -- ([xshift=2.5em]lc3.center) ;
    \draw [thick,->] (ls3.east) -- ([xshift=2.5em]ls3.center) ;
    \draw [thick,->] (fusion.east) -- ([xshift=3em]fusion.west) ;

    \node[](dot1) at ([yshift=-2.5em]lc1.center){$H_f$};
    \node[](dot1) at ([xshift=-1.2em]lc1.center){$L_f$};

    \node[](dot1) at ([yshift=-1.4em]ls1.center){$H_s$};
    \node[](dot1) at ([xshift=-1.7em]ls1.center){$L_s$};


    \draw[decorate,decoration={brace,mirror}] ([xshift=-0.3em,yshift=-1.7em]lc1.center) -- ([xshift=0.3em,yshift=-1.7em]lc1.center);
    
    \draw[decorate,decoration={brace,mirror}] ([xshift=-0.4em,yshift=1.7em]lc1.center) -- ([xshift=-0.4em,yshift=-1.7em]lc1.center);

    \draw[decorate,decoration={brace,mirror}] ([xshift=-1.1em,yshift=-0.8em]ls1.center) -- ([xshift=1.1em,yshift=-0.8em]ls1.center);
    
    \draw[decorate,decoration={brace,mirror}] ([xshift=-1.1em,yshift=0.7em]ls1.center) -- ([xshift=-1.1em,yshift=-0.7em]ls1.center);



    \node[rectangle,minimum height=8.8em,minimum width=19em, very thick, rounded corners, draw,color=gray,fill=gray,opacity=0.1]  (fast_encoder)  at (lc2.center){};

    \node[rectangle,minimum height=6em,minimum width=19em, very thick, rounded corners, draw,color=gray,fill=gray,opacity=0.1]  (slow_encoder)  at (ls2.center){};

    \node[anchor=center,font=\footnotesize] (fast_en) at ([xshift=.4em,yshift=0.5em]lc1.north) {Fast Encoder};
    \node[anchor=center,font=\footnotesize] (fast_en) at ([xshift=.3em,yshift=0.8em]ls1.north) {Slow Encoder};

  \end{tikzpicture}
  \caption{Overview of the proposed TranSFormer, an encoder-decoder paradigm where the encoder involves two separated branches. One is the Slow-branch with subword-level input features and the other is the Fast-Branch with character-level features. $H_f$ is far smaller than $H_s$ for computation efficiency.}
  \label{fig:model-overview}
\end{figure}

Despite the attractiveness of these approaches, previous work is based on an assumption that subwords are the basic units in sequence modeling, and therefore ignores smaller, more fine-grained character-level features. In fact, the benefits of using characters have long been appreciated, and character-based models have been discussed in several sub-fields of NLP, such as language modeling \cite{xue-etal-2022-byt5} and machine translation \cite{lee-etal-2017-fully, li-etal-2021-char, gao-etal-2020-character}. But there are still important problems one needs to address in multi-scale Transformer. The first of these is the computational challenge of dealing with long sequences. For example, when we represent an English text as a character sequence, the length of this sequence is in general 5$\times$ longer than that of the subword sequence. We therefore need to consider this length difference in model design. The second problem is that, from a multiscale learning perspective, learning text representations with features at different levels is not just making use of the syntactic hierarchy of language. To better model the problem, we need some mechanism to describe the interactions among these different linguistic units.

In this study, we aim to exploit the potential of character-level representations in multiscale sequence models while maintaining computational efficiency. Drawing inspiration from the SlowFast convolutional models in video classification~\citep{Christoph2019slowfast}, we propose the Slow-Fast Transformer (TranSFormer) model, which utilizes a fast, thin branch to learn fine-grained character-level features and a slow, wide branch to capture correlations among subword features. A cross-granularity attention layer is placed between the self-attention and feedforward sublayers to make exchanges of cross-granularity information. This enables the slow branch to be aware of fine-grained features while providing optimized high-level representations of the input sequence to the fast branch.

We also make use of character-to-word boundary information to model the interactions among neighboring characters in a word. Additionally, we develop a boundary-wise positional encoding method to better encode the positional information within words for the fast branch. Through a series of extensive experiments on the WMT'14 English-German, WMT'17 Chinese-English and WMT'16 English-Romanian tasks, we demonstrate that TranSFormer yields consistent performance gains while having a negligible increase in the number of parameters and computational cost. As a bonus, our TranSFormer is robust to errors caused by suboptimal tokenization or subword segmentation.

\section{Related Work}
\paragraph{Multiscale Transformer}
Learning multiscale Transformer is a promising way to acquire for further improvements in the machine translation task. A feasible way is to model global and local patterns to enhance Transformer models \cite{shaw-etal-2018-self,yang-etal-2018-modeling,yang-etal-2019-convolutional,Zhao-2019-MUSE}. These work mainly modeled the localness within a fixed window size upon subword input features. Apart from these, \citet{wu-etal-2018-phrase} partitioned the input sequence according to phrase-level prior knowledge, and build attention mechanism upon phrases. Similarly, \citet{hao-etal-2019-multi} proposed a multi-granularity self-attention mechanism, designed to allocate different attention heads to phrases of varying hierarchical structures. Perhaps the most related work to ours is UMST \cite{li2022multiscale}. They re-defined the sub-word, word and phrase scales specific to sequence generation, and modeling the correlations among scales. However, more fine-grained character-level scale is not explored in the previous work due to the serve challenge for encoding long character sequences.

\paragraph{Character-level NMT} 
Fully character-level neural machine translation originates from recurrent machine translation system in \citet{lee-etal-2017-fully}. They built a fully character-level encoder-decoder model, and utilize convolution layers to integrate information among nearby characters. \citet{cherry-etal-2018-revisiting} show the potential of character-level models which can outperform subword-level models under fully optimization. This contributes to their greater flexibility in processing and segmenting the input and output sequences, though modeling such long sequences is time-consuming. More recently, several studies analyze the benefits of character-level systems in multilingual translation scenarios \cite{gao-etal-2020-character}, low-resource translation and translating to typologically diverse languages \cite{li-etal-2021-char}. But these methods all simply view characters as basic units in language hierarchy, and it is still rare to see the effective use of multi-scale learning on character-based language features.

\paragraph{Multi-Branch Transformer}
The utilization of multi-branch architectures has been extensively studied in Transformer models. Early efforts in this area include the Weighted Transformer \cite{Ahmed2017weighted}, which replaced the vanilla self-attention by multiple self-attention branches. Subsequently, the Multi-attentive Transformer \citep{Fan-multibranch-attentive} and Multi-Unit Transformer \citep{yan-etal-2020-multi} have advanced this design schema by incorporating branch-dropout and switching noise inputs, respectively. Additionally, \citet{wu2020mixed} investigated the potential advantages of utilizing dual cross-attention mechanisms to simultaneously attend to both Sentencepiece \cite{kudo-richardson-2018-sentencepiece} and subword \cite{sennrich-subword-neural}.
In this work, we take a forward step to exploit the potential of character features. We argue that a lightweight branch is sufficient to encode useful fine-grained features, an aspect that has not been previously investigated.
\input{./Figure/char_matrix.tex}

\section{Method}
The proposed TranSFormer follows a encoder-decoder paradigm (see Figure \ref{fig:model-overview}) which involves two encoder branches operating at different input granularities. The original subword encoder, which has a large model capacity for fully learning correlations among input individuals, is defined as the slow branch. The other branch, designed to handle character-level representations using a thin encoder to efficiently capture correlations among characters, is referred to as the fast branch. Our goal is to use the fast branch to learn a fine-grained but less precise representation to complement the slow branch. In the following sections, we will elaborate the core design of Slow branch, Fast branch and the cross-granularity attention, respectively.

\subsection{The Slow Branch for Subwords}
We use the standard Transformer as the slow branch due to its strong ability to model global interactions among input sequences. The input of the slow branch is the mixture of subwords and words since some high-frequency words have not been further divided into subwords. Following the suggestions in \citet{li2022multiscale}, we adopt a graph convolutional network to model the inner correlations among words through the adjacency matrix $\mathcal{A}_{s}$. To this end, the Slow branch then encodes the enhanced representation via the self-attention mechanism, $ \mathrm{SAN} = \mathrm{Softmax}(\frac{Q \cdot K^{\mathsf{T}}}{\sqrt{d_k}}) \cdot V$, where $Q$, $K$, $V$ are obtained through three independent projection matrix, such as $W_q$, $W_k$, $W_v$. A point-wise feed-forward network is followed, $\mathrm{FFN} = \textrm{\texttt{max}}(xW_{1} + b_1, 0)W_2 + b_2$, where $W_1$ and $W_2$ are transformation matrices and $b_1$ and $b_2$ are bias matrices. To bridge the gap between two granularities, we sandwich a new sublayer between the self-attention and the feed-forward network, to accomplish the feature interaction between the slow and fast branches. A straightforward idea is to employ a cross-attention similar with encoder-decoder attention in the decoder side. We will discuss more details in the Section \ref{sec:fusion}.

\input{Figure/arch_head_ds_ln.tex}
\subsection{The Fast Branch for Characters}
To enhance the efficiency of modeling long character-level inputs, we propose the use of a fast branch with a tiny hidden size. The hidden size is a critical factor in the computation of the self-attention network \cite{vaswani2017attention}, and by reducing it, we can achieve faster computation. To the best of our knowledge, this is the first attempt to design multiscale Transformer models that considers character-level features, as the long input sequence has previously hindered such exploration. 

While the fast branch may not be as powerful as the slow branch, it is still effective in learning fine-grained features. Our initial experiments have yielded two notable findings: 1) a slow branch with hidden size of 32 is sufficient for transferring fine-grained knowledge to the slow branch, and 2) cross-granularity fusion is crucial for the slow branch, while removing the reversed fusion in the fast branch has only a moderate effect on performance. We would ablate this settings in the Section \ref{sec:results}. To further improve the modeling ability, we introduce several techniques as follows:

\paragraph{Char Boundary Information} 
The use of word-boundary information has been shown to effectively reduce the redundant correlations among subwords, as demonstrated in \cite{li2022multiscale}. This leads to the consideration of character-level modeling, which poses a more challenging problem due to the greater number of characters typically present within a word in comparison to subwords. The statistical analysis in Figure \ref{fig:distribution} further evidences it that a significant proportion of words contain more than 5 characters, while a much smaller number are divided into subwords. Thus, model may be unable to discern the distinction between the same character that belongs to the same word and that of distinct words.

To address this issue, we propose the use of a character-level graph convolution network (GCN) to learn local, fine-grained features while also allowing each character to be aware of its proximity to other characters. GCN\cite{Kipf2017gcn} is a suitable choice for this task as it aggregates feature information from the neighbors of each node to encapsulate the hidden representation of that node. The computation can be described as:
\begin{equation} \label{eq:gcn-char}
      \mathrm{GCN}_{\mathrm{Fast}} = \sigma(\tilde{D}_f^{-\frac{1}{2}} \tilde{\mathcal{A}_{f}} \tilde{D}_f^{-\frac{1}{2}} \cdot x W_f^g),
    \end{equation}
\noindent $\tilde{\mathcal{A}_{f}}={\mathcal{A}_{f}}+\mathcal{I}_{L}$ denotes the adjacency matrix of the undirected graph with self-connections. Here $\mathcal{I}_L$ denotes the identity matrix. $\tilde{D}_f$ is the degree matrix of the adjacency matrix $\tilde{\mathcal{A}_f}$. $W_f^g$ is a linear transformation which is a trainable parameter. The character-level encoder architecture is illustrated in Figure \ref{fig:architecture}. 

\begin{figure}[t]
    \centering
    \hspace{-2em}
    \subfloat[Relative-positional-representations]{
    \label{fig:char2word_matrix0} 
    \begin{tikzpicture}[]
        \tikzstyle{every node}=[scale=0.4]
        \tikzstyle{node1} = [anchor=center,draw,minimum height=2em,minimum width=2em,inner sep=0pt,fill=black!60]
        \tikzstyle{node2} = [anchor=center,draw,minimum height=2em,minimum width=2em,inner sep=0pt,fill=ugreen!80]
        \tikzstyle{node3} = [anchor=center,draw,minimum height=2em,minimum width=2em,inner sep=0pt,fill=ugreen!50]
        \tikzstyle{node4} = [anchor=center,draw,minimum height=2em,minimum width=2em,inner sep=0pt,fill=ugreen!20]
        
        \tikzstyle{node5} = [anchor=center,draw,minimum height=2em,minimum width=2em,inner sep=0pt]
        
        \tikzstyle{node6} = [anchor=center,draw,minimum height=2em,minimum width=2em,inner sep=0pt,fill=yellow!20]
        \tikzstyle{node7} = [anchor=center,draw,minimum height=2em,minimum width=2em,inner sep=0pt,fill=yellow!50]
        \tikzstyle{node8} = [anchor=center,draw,minimum height=2em,minimum width=2em,inner sep=0pt,fill=yellow!80]
        \tikzstyle{node9} = [anchor=center,draw,minimum height=2em,minimum width=2em,inner sep=0pt,fill=black!60]
        
        \foreach \i / \j / \k / \z in {0/8/5/0, 1/8/6/1, 2/8/7/2, 3/8/8/3, 4/8/9/4, 5/8/9/4, 6/8/9/4, 7/8/9/4, 8/8/9/4, 0/7/4/-1, 1/7/5/0, 2/7/6/1, 3/7/7/2, 4/7/8/3, 5/7/9/4, 6/7/9/4, 7/7/9/4, 8/7/9/4, 0/6/3/-2, 1/6/4/-1, 2/6/5/0, 3/6/6/1, 4/6/7/2, 5/6/8/3, 6/6/9/4, 7/6/9/4, 8/6/9/4, 0/5/2/-3, 1/5/3/-2, 2/5/4/-1, 3/5/5/0, 4/5/6/1, 5/5/7/2, 6/5/8/3, 7/5/9/4, 8/5/9/4, 0/4/1/-4, 1/4/2/-3, 2/4/3/-2, 3/4/4/-1, 4/4/5/0, 5/4/6/1, 6/4/7/2, 7/4/8/3, 8/4/9/4, 0/3/1/-4, 1/3/1/-4, 2/3/2/-3, 3/3/3/-2, 4/3/4/-1, 5/3/5/0, 6/3/6/1, 7/3/7/3, 8/3/8/3, 0/2/1/-4, 1/2/1/-4, 2/2/1/-4, 3/2/2/-3, 4/2/3/-2, 5/2/4/-1, 6/2/5/0, 7/2/6/2, 8/2/7/2, 0/1/1/-4, 1/1/1/-4, 2/1/1/-4, 3/1/1/-4, 4/1/2/-3, 5/1/3/-2, 6/1/4/-1, 7/1/5/0, 8/1/6/1, 0/0/1/-4, 1/0/1/1, 2/0/1/-4, 3/0/1/-4, 4/0/1/-4, 5/0/2/-3, 6/0/3/-2, 7/0/4/-1, 8/0/5/0} \node[node\k] (a\i\j) at (0.8em*\i + 0*0em, 0.8em*\j + 0*0em) {};
        
        \foreach \i / \j / \k in {0/8/g, 0/7/o, 0/6/o, 0/5/d, 0/4/-, 0/3/n, 0/2/e, 0/1/w, 0/0/s} 
        \node[anchor=center] (n\k\j) at ([xshift=-0.5em,yshift=0em]a\i\j.west) {\LARGE \k};
        
        \foreach \i / \j / \k in {0/0/g, 1/0/o, 2/0/o, 3/0/d, 4/0/-, 5/0/n, 6/0/e, 7/0/w, 8/0/s} 
        \node[anchor=center] (n\k\i) at ([xshift=-0em,yshift=-0.5em]a\i\j.south) {\LARGE \k};
        
    \end{tikzpicture}
    }
    \hspace{1em}
    \subfloat[Boundary-wised positional encoding]{
    \label{fig:char2word_matrix1} 
    \begin{tikzpicture}

         \tikzstyle{every node}=[scale=0.4]
        \tikzstyle{node1} = [anchor=center,draw,minimum height=2em,minimum width=2em,inner sep=0pt,fill=black!60]
        \tikzstyle{node2} = [anchor=center,draw,minimum height=2em,minimum width=2em,inner sep=0pt,fill=ugreen!80]
        \tikzstyle{node3} = [anchor=center,draw,minimum height=2em,minimum width=2em,inner sep=0pt,fill=ugreen!50]
        \tikzstyle{node4} = [anchor=center,draw,minimum height=2em,minimum width=2em,inner sep=0pt,fill=ugreen!20]
        
        \tikzstyle{node5} = [anchor=center,draw,minimum height=2em,minimum width=2em,inner sep=0pt]
        
        \tikzstyle{node6} = [anchor=center,draw,minimum height=2em,minimum width=2em,inner sep=0pt,fill=yellow!20]
        \tikzstyle{node7} = [anchor=center,draw,minimum height=2em,minimum width=2em,inner sep=0pt,fill=yellow!50]
        \tikzstyle{node8} = [anchor=center,draw,minimum height=2em,minimum width=2em,inner sep=0pt,fill=yellow!80]
        \tikzstyle{node9} = [anchor=center,draw,minimum height=2em,minimum width=2em,inner sep=0pt,fill=black!60]
        
        \begin{scope}[scale=1]
        \foreach \i / \j / \k / \z in
            {0/8/5/0, 1/8/6/1, 2/8/7/2, 3/8/8/3, 4/8/9/4, 5/8/9/4, 6/8/9/4, 7/8/9/4, 8/8/9/4,
            0/7/4/-1, 1/7/5/0, 2/7/6/1, 3/7/7/2, 4/7/9/4, 5/7/9/4, 6/7/9/4, 7/7/9/4, 8/7/9/4,
            0/6/3/-2, 1/6/4/-1, 2/6/5/0, 3/6/6/1, 4/6/9/4, 5/6/9/4, 6/6/9/4, 7/6/9/4, 8/6/9/4,
            0/5/2/-3, 1/5/3/-2, 2/5/4/-1, 3/5/5/0, 4/5/9/4, 5/5/9/4, 6/5/9/4, 7/5/9/4, 8/5/9/4,
            0/4/1/4, 1/4/1/4, 2/4/1/4, 3/4/1/4, 4/4/5/0, 5/4/9/4, 6/4/9/4, 7/4/9/4, 8/4/9/4,
            0/3/1/4, 1/3/1/4, 2/3/1/4, 3/3/1/4, 4/3/1/4, 5/3/5/0, 6/3/6/1, 7/3/7/2, 8/3/8/3,
            0/2/1/4, 1/2/1/4, 2/2/1/4, 3/2/1/4, 4/2/9/4, 5/2/4/-1, 6/2/5/0, 7/2/6/1, 8/2/7/2,
            0/1/1/4, 1/1/1/4, 2/1/1/4, 3/1/1/4, 4/1/9/4, 5/1/3/-2, 6/1/4/-1, 7/1/5/0, 8/1/6/1,
            0/0/1/4, 1/0/1/4, 2/0/1/4, 3/0/1/4, 4/0/9/4, 5/0/2/-3, 6/0/3/-2, 7/0/4/-1, 8/0/5/0}
            \node[node\k] (a\i\j) at (0.8em*\i + 0*0em,0.8em*\j + 0*0em) {};
       
        \foreach \i / \j / \k in
            {0/8/g, 0/7/o, 0/6/o, 0/5/d, 0/4/-, 0/3/n, 0/2/e, 0/1/w, 0/0/s}
            \node[anchor=center] (n\k\j) at ([xshift=-0.5em,yshift=0em]a\i\j.west) {\LARGE \k};
         
        \foreach \i / \j / \k in
            {0/0/g, 1/0/o, 2/0/o, 3/0/d, 4/0/-, 5/0/n, 6/0/e, 7/0/w, 8/0/s}
            \node[anchor=center] (n\k\i) at ([xshift=-0em,yshift=-0.5em]a\i\j.south) {\LARGE \k};

        \end{scope}
  
    \end{tikzpicture}
    }
    \caption{Comparison of the relative positional encoding (a) and our proposed boundary-wised positional encoding method (b). Note that characters in a dark color means the mask for exceeding the max window. }
    \label{fig:relative_pos_emb}
\end{figure}
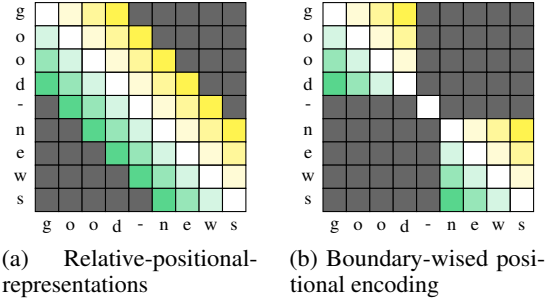

\paragraph{Boundary-wised Positional Encoding}
To further enhance the relative positional representation among characters, we design a boundary-wised positional encoding (PE) method. Our intuition is to provide each character with the ability to recognize which characters belong to the same word. Thus we restrict the relative window within each word, as illustrated in Figure \ref{fig:relative_pos_emb}. Vanilla relative positional encoding \cite{shaw-etal-2018-self} models the correlations in a fixed window size $2k + 1$. Here we set $k=3$, positions exceed $k$ would be masked. Differently, the proposed boundary-wised PE is utilized to enhance the inner relative positional information among characters within each word. In our preliminary experiments, boundary-wised PE is helpful for stable training.

\subsection{Cross-Granularity Fusion}
\label{sec:fusion}
As depicted in Figure \ref{fig:architecture}, the computation of the two branches in our TranSFormer architecture is separated, with each branch operating independently of the other's representation. To facilitate communication between the branches, we propose the utilization of a cross-granularity information fusion method within each encoder block. This method can be implemented through various options. Given the lengths of the slow and fast branches as $L_s$ and $L_f$, and the hidden sizes as $H_s$ and $H_f$, respectively, the goal is to seamlessly integrate cross-scale information within each encoder block between the two branches. In the field of machine translation, it is straightforward to employ cross-attention mechanisms, such as encoder-decoder attention \cite{vaswani2017attention} or context-aware cross-attention \cite{voita-etal-2018-context}, to capture correlations between the representations. 

Our default strategy is to employ a cross-granularity attention mechanism (namely CGA) sandwiched between the self-attention and feedforward network. The architecture is plotted in Figure \ref{fig:architecture}. $x_f$ and $x_s$ denote the representation of the fast and slow branches, respectively. The challenge remains here is the mismatched feature shape between $x_s$ and $x_f$  Here, we take the Fast branch as an instance, we first normalize $x_s$ via $\hat{x}_{f} = \mathrm{LN}(x_s)$. $\mathrm{LN(\cdot)}$ denotes the layer normalization for stable optimization. Then $\hat{x}_{f}$ is fed into CGA of the fast branch, the formulation is as follows:
\begin{eqnarray}
    \textrm{ATTN}_{f} &=& \mathrm{Softmax}(\frac{x_f W_f^{q} \cdot ({\hat{x}_f}W_f^{k})^{\mathsf{T}}}{\sqrt{d_f^k}}),  \nonumber \\
    \textrm{CGA} &=& \textrm{ATTN}_{f} \cdot {\hat{x}_f W_f^{v}}, \label{eq:cross-view-attn}
\end{eqnarray}
\noindent where the query is derived from the residual output of SAN in the Fast branch via $x_s \cdot W_f^{q}$. The key and value are derived from the Slow branch via ${\hat{x}_f}W_f^{k}$ and ${\hat{x}_f}W_f^{v}$, respectively. It is worthy to note that, the shape of $W_f^k$ and $W_f^v$ $\in \mathbb{R}^{H_s \times H_f}$, to reduce the hidden size. Detailed transformation could be found in the left part of Figure \ref{fig:architecture}.

It is important to note that our proposed method of cross-granularity fusion is bidirectional, as opposed to the lateral connections used in the SlowFast \cite{Christoph2019slowfast}. Other alternative methods would be discussed in Section \ref{sec:analysis}.

\subsection{Interactions Between Encoder and Decoder}
\label{sec:interaction}
In vanilla Transformer, the key and value of the encoder-decoder attention on the decoder side derives from the encoder output, however, there are two branches in our TranSFormer (See Figure \ref{fig:model-overview}). It is worthy to investigate how to effectively leverage the multi-granularity representations. Our default strategy is to regard the fast branch as an auxiliary to provide fine-grained features for the slow branch, thus only the output of the slow branch is exposed to the decoder. Besides this, there are also several feasible options. For example, we can fuse the outputs of two branches as the final encoder output, or building a double-branch encoder-decoder attention to attend two branches independently. We compares this options in our experiments.

\begin{table*}[t!]
    \centering
    \small
    \setlength{\tabcolsep}{3.5pt}
    \begin{tabular}{llllrlrl}
    \Xhline{1.0pt}
        \multicolumn{2}{c}{\multirow{2}{*}{\textbf{Model}}} &  \multirow{2}{*}{\textbf{Enc.}} & \multirow{2}{*}{\textbf{Dec.}} & \multicolumn{2}{c}{\textbf{Base}} & \multicolumn{2}{c}{\textbf{Big}}\\
        & & & & \bf Param  & \bf  BLEU  & \bf Param  & \bf BLEU \\
        \hline
        & Transformer~\cite{vaswani2017attention}          &  Sub       &  Sub &  65M  &  27.30  & 213M & 28.40  \\
        & Transformer                                      &  Char      &  Sub &  63M  &  26.56  & 208M & 28.05  \\
        \hline
        \multirow{5}{*}{\textbf{Multiscale}}
        & RPR~\cite{shaw-etal-2018-self}                   &  Sub       &  Sub &  65M  &  27.60  & 213M & 29.20  \\
        & CSAN~\cite{yang-etal-2019-convolutional}         &  Sub       &  Sub &  88M  &  28.18  & -    & 28.74  \\
        & Localness~\cite{yang-etal-2018-modeling}         &  Sub       &  Sub &  89M  &  28.11  & 267M & 29.18  \\
        & \textsc{Mg-Sa}~\cite{hao-etal-2019-multi}        &  Sub       &  Sub &  89M  &  28.28  & 271M & 29.01  \\
        & UMST~\cite{li2022multiscale}                     &  Sub       &  Sub &  70M  &  28.51  & 242M & 29.75  \\
        \hline
        \multirow{3}{*}{\textbf{Double-Branch}}
        & Muse~\cite{Zhao-2019-MUSE}                       &  Sub       &  Sub &  -    &   -     & 233M  &29.90  \\
        & Multi-Attentive~\cite{Fan-multibranch-attentive} &  Sub       &  Sub &  -    &   -     & 325M& 29.80  \\

        & Multi-Unit~\cite{yan-etal-2020-multi}            &  Sub       &  Sub &  130M &  29.30  & -    & -  \\
        \hline
        \multirow{3}{*}{\textbf{Character-level}}
        & ConvTransformer~\cite{gao-etal-2020-character} $\dag$ &  Char &  Char&  65M &  23.47  &-& -  \\   
        & Fast Only (Hidden=32, L=6)                       &  Char      &  Sub &  42M   &  17.90$_{(16.9)}$ & -&  -      \\
        & Fast Only (Hidden=512, L=6)                      &  Char      &  Sub &  64M  &  27.11$_{(26.1)}$  & 211M & 28.65$_{(27.6)}$ \\   
        \hline
        \multirow{3}{*}{\textbf{Slow-Fast}}
        & Slow Only                                        &  Sub       &  Sub &  63M  &  27.40$_{(26.4)}$  & 211M& 28.80$_{(27.8)}$  \\   
        & TranSFormer  (Hidden=32, L=6)                    &  Char/Sub  &  Sub &  66M  &  28.56$_{(27.6)}$  & 231M &29.85$_{(28.9}$  \\
        & TranSFormer + ODE \cite{li-etal-2022-ode}        &  Char/Sub  &  Sub &  66M  &  29.30$_{(28.3)}$  & - & - \\
        \hline
    \end{tabular}
    \caption{Comparison with previous studies on the WMT En-De task. Models with $\dag$ denote the re-implementing results based on our codebase within the same hyperparameters. BLEU at the right corner denotes the SacreBLEU.}
\label{tab:main-results}
\end{table*}

\begin{table}[t]
  \setlength{\tabcolsep}{2.0pt}
  \small
  \centering
  \begin{tabular}{lrr}
  \Xhline{1.0pt}
  \multicolumn{3}{c}{\textbf{(a) Previous work based on Big models}} \\
  \bf System & \bf Params &  \bf BLEU \\
  \hline
  Transformer-Big\citep{hassan2018achieving}            &-         & 24.20 \\
  CSAN~\cite{yang-etal-2019-convolutional}              &-         & 25.01 \\
  Localness~\cite{yang-etal-2018-modeling}              & 307M     & 25.03 \\
  \textsc{Umst}~\cite{li2022multiscale}                 & 307M     & 25.23 \\
  \Xhline{1.0pt}
  \multicolumn{3}{c}{\textbf{(b)  Our Big models}} \\
  subword-level Transformer-Big                         & 261M     & 24.41 \\
  character-level Transformer-Big                       & 258M       & 23.80 \\

  TranSFormer (Hidden=64)                         & 283M     & 25.55 \\
  \hline
  \end{tabular}
  \caption{Results on WMT Zh-En. We compare several prior work of learning local patterns.}
  \label{tab:wmt-zh2en}
\end{table}

\section{Experiments}
\subsection{Experimental Setups}
\paragraph{Datasets}
 The present study examines the performance of our proposed TranSFormer on several machine translation datasets: the WMT'14 English-German (En-De), WMT'16 English-Romanian (En-Ro) and WMT'17 Chinese-English (Zh-En) datasets. The En-De dataset comprises approximately $4.5$ million tokenized sentence pairs, which were preprocessed following the same procedure as in \citet{ott2018scaling} to yield a high-quality bilingual training dataset. For validation, we use the \textit{newstest2016} set, while the \textit{newstest2014} set served as the test data. The En-Ro dataset consists of $610$K bilingual sentence pairs, and we adopt the same preprocessing scripts as in \citet{lee-etal-2018-deterministic,kasai2020disco}, using a joint source and target BPE factorization with a vocabulary size of $40$K. The \textit{newsdev2016} set is used for validation, while the \textit{newstest2016} set served as the test set. For the Zh-En task, we collect all the available parallel data for the WMT17 Chinese-English translation task, consisting $15.8$M sentence pairs from the UN Parallel Corpus, $9$M sentence pairs from the CWMT Corpus and about $332$K sentence pairs from the News Commentary corpus. After carefully data filtering setups in \citet{hassan2018achieving}, there are left $18$M bilingual pairs. \textit{newsdev2017} and \textit{newstest2017} are served as the validation and test sets, respectively.

\paragraph{Setups} 
For the machine translation task, we mainly evaluate the proposed TranSFormer on base and big configurations. The hidden size of slow branch is $512$/$1024$ for base and big, respectively. And the filter size in FFN is $2048$/$4096$. In our default setting, a width of $32$ slow branch is enough to learn fine-grained features, and the corresponding filter size is set to $128$. We both employ residual dropout, attention dropout and activation dropout. All values are $0.1$, except the residual dropout $0.3$ for big counterparts.

\paragraph{Training and Evaluations}
The codebase is developed upon \textit{Fairseq} \cite{ott2019fairseq}. All experiments are conducted on 8 Tesla V100 GPUs. We use Adam \cite{kingma2014adam} optimizer with ($0.9$, $0.997$), and the default learning rate schedule with $0.002$ max value, $16,000$ warmup steps. For machine translation tasks, BLEU scores are computed by \textit{mult-bleu.perl}, and we also provide the SacreBLEU\footnote{BLEU+case.mixed+numrefs.1+smooth.exp+ tok.13a+version.1.2.12} for En-De. The beam size is $4$ for En-De and $8$ for Zh-en, and the length penalty is $0.6$ and $1.3$, respectively.

    \begin{table}[t]
      \small
      \setlength{\tabcolsep}{5.0pt}
      \centering
      \begin{tabular}{lrr}
      \Xhline{1.0pt}
      \textbf{Model} & \bf Param & \bf BLEU \\
      \hline
      \textsc{Delight} \cite{mehta2020delight}                                  & 53M   &  34.70     \\
      Baseline in \textsc{mBART} \cite{liu-etal-2020-multilingual-denoising}    & -     &  34.30     \\
      Baseline in \textsc{Disco} \cite{kasai2020disco}                          & -     &  34.16     \\
      Transformer $\dag$ \cite{vaswani2017attention}                            & 54M   &  34.21     \\
      \textsc{TNT}$ \dag$  \cite{han2021transformer}                            & 73M   &  34.00     \\
      \textsc{Umst}~\cite{li2022multiscale}                                     & 60M   &  34.81     \\
      ODE Transformer~\cite{li-etal-2022-ode}                                   & 69M   &  34.94     \\
      \hline
      TranSFormer (Hidden=32)                                                   & 59M  & \bf 35.40  \\
      \hline
    \end{tabular}
    \caption{Results on the WMT En-Ro task.}
    \label{tab:en-ro}
    \end{table}

    \begin{table*}[t]\centering\vspace{-1em}
    \captionsetup[subfloat]{captionskip=2pt}
    \small
    \subfloat[\textbf{Char-width}: Comparisons of various char width on performance.
    \label{tab:char_width}]{
    \setlength{\tabcolsep}{2.0pt}
        \begin{tabular}{l|c|cc|cc}
          \multicolumn{1}{c|}{Model} & Width & En-De & Param. & Zh-En & Param. \\
          \Xhline{1.0pt}
          Transformer          & -    & 27.20 & 63.1M  & 24.41 & 261.9M \\
          \hline
          TranSFormer          & 16   & 28.10 & 66.6M  & 25.20 & 281.0M  \\
          TranSFormer          & 32   & 28.56 & 66.9M  & 25.47 & 281.7M  \\
          TranSFormer          & 64   & 28.39 & 67.6M  & 25.55 & 283.1M  \\
          TranSFormer          & 128  & 28.27 & 69.7M  & 25.51 & 286.6M  \\
          TranSFormer          & 256  & 28.17 & 76.1M  & 25.33 & 295.7M  \\
          \end{tabular}}
          \hspace{2em}
    \subfloat[\textbf{Fusion Method}: Fusing Slow and Fast branches with several types of methods.
    \label{tab:fusion}]{
        \begin{tabular}{l|cc}
          \multicolumn{1}{c|}{Model} & Fusion Methods & BLEU \\
          \Xhline{1.0pt}
          Slow only      & -    & 27.20 \\
          Fast only      & -    & 17.90 \\      
          \hline
          TranSFormer           & CGA  & 28.56 \\
          TranSFormer           & Linear + Attention  & 28.47 \\
          TranSFormer           & DS + Concat  & 27.82 \\
          TranSFormer           & DS + Sum     & 27.96 \\
          
        \end{tabular}}
        \hspace{2em}
    \subfloat[\textbf{Interactions}: Figuring out the impact of various encoder-decoder interaction manners on performance.
    \label{tab:interactions}]{
        \begin{tabular}{l|cc}
          \multicolumn{1}{c|}{Model} & Enc. Output & BLEU \\
          \Xhline{1.0pt}
          TranSFormer                   & Slow Branch          & 28.56 \\
          TranSFormer                   & Fast Branch          & 23.11 \\
          TranSFormer                   & Both Slow and Fast     & 28.25 \\
          \end{tabular}}
    \hspace{2em}
    \subfloat[\textbf{Input}: Figuring out the impact of various input granularites for Slow and Fast branch.
    \label{tab:input}]{
        \begin{tabular}{l|c|c|c}
          \multicolumn{1}{c|}{\multirow{2}*{Model}} & \multicolumn{2}{c|}{Input Granularity} & \multirow{2}*{BLEU} \\
          \cline{2-3}
          \multicolumn{1}{c|}{} & Slow-Branch & Fast-Branch \\
          \Xhline{1.0pt}
          TranSFormer                   & Subword & Character       & 28.56 \\
          TranSFormer                   & Subword & Subword         & 27.50 \\
          TranSFormer                   & Subword & Sentencepiece   & 28.27 \\
          TranSFormer                   & Sentencepiece &  Character  & 28.60 \\
          \end{tabular}}
          \caption{Ablations on TranSFormer design on the WMT En-De task. The evaluation metric is BLEU (\%). We mainly ablate the experiments from the width of fast branch, various fusion methods, the interactions between encoder-decoder, and the input granularity.}
    \label{tab:ablations}
\end{table*}
\paragraph{Results of En-De}
The results of the WMT En-De task under both base and big configurations are summarized in Table \ref{tab:main-results}. As evidenced by the results, our TranSFormer model demonstrates significant improvements in BLEU when compared to the Slow only model, with gains of up to $1.16$/$1.05$ BLEU scores under the base/big configurations. Conversely, the Fast only baseline, which has a hidden size of $32$, only attains a BLEU score of $17.90$, leading to a considerable performance gap due to its limited capacity. However, it still contributes up to a $1.14$ BLEU-point benefit to the Slow branch, indicating that the fine-grained correlations modeled by the Fast branch are complementary.  Additionally, we present the results of prior works, which have employed both character-level and subword-level systems, and categorize them in terms of various aspects. TranSFormer can beat or on par with prior works with less parameters. This indicates the investigation of character-level mutliscale models is meaningful. Note that TranSFormer is computationally efficient, only requiring additional $15\%$ training cost and negligible inference latency. And TranSFormer can also benefit from advanced design, \textit{e.g.}, another $0.74$ improvement with ODE method \cite{li-etal-2022-ode}.

\subsection{Results}
\label{sec:results}

\paragraph{Results of Zh-En}
 The WMT'17 Zh-En task poses a significant challenge due to the linguistic differences between Chinese and English. Additionally, the Chinese language owns less characters per word than English.   Table \ref{tab:wmt-zh2en} shows the results of our comparison of the TranSFormer with prior works. We observe TranSFormer yields a $1$ BLEU point improvements than the subword-level systems. Our TranSFormer model demonstrates superior performance compared to previous work that models local patterns, while maintaining efficient computational requirements. We will exploit whether TranSFormer can gain more benefits when incorporating these techniques on the slow branch.

\paragraph{Results of En-Ro}
Furthermore, our empirical evaluations of the proposed TranSFormer architecture on the smaller WMT En-Ro dataset also demonstrate consistent improvements in BLEU scores as a result of the utilization of interactions among granularities. Notably, the TranSFormer model even outperforms the ODE Transformer \cite{li-etal-2022-ode}, an advanced variant that leverages the advantages of high-order ordinary differential equations (ODE) solutions, by a substantial margin while incurring less computational cost.

\subsection{Analysis}
\label{sec:analysis}
This section provides ablation studies of TranSFormer in terms of several core techniques.

\paragraph{Effect of width on Fast branch }
We first aim to explore TranSFormer under various widths of the Fast branch, including $16$, $32$, $64$, $128$ and $256$. Results in Table \ref{tab:char_width} show that even a hidden size of $16$ can provide helpful fine-grained features for the slow branch, and yielding almost $1$ BLEU-point gains by bringing modest parameters. Empirically, a hidden of $32$ and $64$ deliver the best performance on base (En-De) and big (Zh-En) configurations, respectively. Further increasing the hidden layer dimension of the model results in no more gains, while requiring more computational cost.

\paragraph{Fusion methods between branches}
In addition to our proposed fusion method CGA, there are several alternative techniques that can be considered. The most straightfoward one is to transform the hidden with a linear projection and then use a standard cross-attention. It delivers similar performance but consumes more parameters. Another option is to downsample the character-level representation from $L_f$ to $L_s$, and then concatenate (namely DS + Concat) or sum (DS + Sum) the two representations. Although both of these methods have been found to outperform the Slow only baseline, they have not been found to be on par with CGA method. This may be due to the fact that downsampling may impede optimization due to the low compression ratio of text compared with images.

\paragraph{Various interaction methods}
In Table \ref{tab:interactions}, we present a summary of various promising options for interactions between the encoder and decoder. Empirical results indicate that utilizing the Slow branch as the output yields the highest performance. This can be attributed to the Fast branch's ability to provide fine-grained features and low-level semantic knowledge as auxiliary information to the Slow branch. Additionally, while utilizing the Fast branch as the encoder output results in inferior performance compared to the baseline, it still yields a significant improvement over the Slow only baseline ($17.90$). This highlights the effectiveness of the TranSFormer model in leveraging interactions between different granularities. Furthermore, we also evaluated a two-stream approach in the decoder, in which one stream attends to the Slow branch and the other attends to the Fast branch, with a gated mechanism being used to fuse the features. However, this method was not sufficient to further improve performance. We attribute this to the negative interactions brought by the Fast branch, increasing the optimization difficulty.
\begin{table}
    \centering
    \small
    \begin{tabular}{l|l|l}
        \multicolumn{1}{c|}{\#} & \multicolumn{1}{c|}{Model} & BLEU \\
        \Xhline{1.0pt}
        1& TranSFormer & 28.56 \\
        2&  \;w/ unidirectional CGA          & 28.41 \\
        3&  \;w/o character-boundary                   & 27.61 \\
        4&  \;w/ replace boundary with random          & 26.79 \\
        5&  \;w/o boundary-wised PE               & 28.13 \\
        6&  \;w/ linear attention \cite{wang2020linformer}    & 28.08 \\
    \end{tabular}
    \caption{\label{tab:ablation_study}
    Ablations on the fast branch in terms of several core design schemas.
    }
\end{table}

\begin{table}
    \centering
    \setlength{\tabcolsep}{6.0pt}
    \small
    \begin{tabular}{l|l|l}
        \multicolumn{1}{c|}{\#} & \multicolumn{1}{c|}{Model}  & BLEU  \\
        \Xhline{1.0pt}
        1 & TranSFormer (default: all blocks)   &  28.56 \\
        2 & \;+ at the last encoder block (\textit{e.g.}, 6)  & 27.99 \\
        3 & \;+ at bottom 3 blocks (\textit{e.g.}, 1/2/3)    & 28.10 \\
        4 & \;+ at top 3 blocks (\textit{e.g.}, 4/5/6) & 28.40 \\
        5 & \;+ every 2 blocks (\textit{e.g.}, 1/3/5) & 28.20 \\
    \end{tabular}
    \caption{
    Ablations on operating interactions between two branches in different levels.}
    \label{tab:ablation_fusion_methods}
\end{table}

\paragraph{Effect of various input granularities}
To ascertain whether the observed performance gains can be attributed to the complementary information provided by fine-grained character-level representations, we replaced the input of the fast branch with subword-level sequences, identical to that of the slow branch. The results presented in Table \ref{tab:input} demonstrate a degradation of up to $1$ BLEU point. This can be attributed to the lack of distinct or complementary features provided by the fast branch and the limited capacity of the model in fully optimizing subword-level features. This observation further supports the hypothesis that the Slow-Fast design can learn complementary features for each granularity. Furthermore, we found that the TranSFormer architecture with sentencepiece \cite{kudo-richardson-2018-sentencepiece} as the fast branch input can also benefit from the two-branch design, due to the different segmentations. Additionally, our TranSFormer is a general design that can work well with a character-level fast branch and a sentencepiece-level slow branch, yielding a BLEU score of $28.60$, even slightly better than the subword-level one.

\paragraph{Ablations on fast branch designs}
It is hard to directly learn the tedious character sequence. The proposed character-boundary injection serves as a crucial component in addressing this challenge. Without this injection, the TranSFormer model suffers from a significant decrease in BLEU (\#3). Furthermore, the situation is exacerbated when the boundary is replaced with a randomly initialized one (\#4), emphasizing the importance of the proposed character-boundary injection. Also, both removing the boundary-wised positional encoding (\#5) or replacing the vanilla attention by linear attention \cite{wang2020linformer} (\#6) lead to modest BLEU degradation. While, there is no significant impact when using unidirectional CGA (\# 2, from Fast to Slow).

\begin{table}
    
    \centering
    \small
    \begin{tabular}{l|l|l|l|l}
        \multicolumn{1}{c|}{\#} & Model & 50K &500K & 1000K \\
        \Xhline{1.0pt}
        1 & TranSFormer     & 11.87 & 22.75 & 25.30 \\
        2 & Character-only  & 10.50  & 20.50  & 22.50 \\
        3 & Subword-only    & 7.00  & 22.00 & 23.50 \\
    \end{tabular}
    \caption{Comparison of different low-resource settings, including 50K, 500K, and 1000K training subsets sampled from the WMT En-De dataset.}
    \label{tab:low_resource}
\end{table}

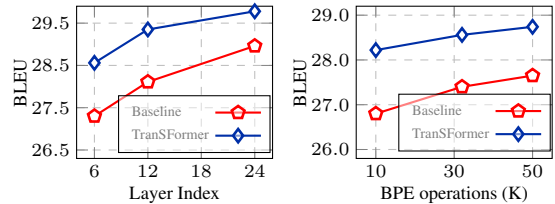
\begin{figure}
    \centering
    \vspace{-0.8em}
    \subfloat {
    \begin{tikzpicture}[]
      \scriptsize{
      \begin{axis}[
       at={(0,0)},
        ymajorgrids,
        xmajorgrids,
        grid style=dashed,
        width=.26\textwidth,
        height=.225\textwidth,
        legend style={at={(0.45,0.14)}, anchor=south west},
        xlabel={\scriptsize{Layer Index}},
        ylabel={\scriptsize{BLEU}},
        ylabel style={yshift=-2em},xlabel style={yshift=1.0em},
        yticklabel style={/pgf/number format/precision=1,/pgf/number format/fixed zerofill},
        ymin=26.30,ymax=29.90, ytick={26.50, 27.50, 28.50, 29.50},
        xmin=4,xmax=26,xtick={6,12,18,24},
        legend style={yshift=-6pt,xshift=-2.5em, legend plot pos=right,fill opacity=0.5,font={\tiny},cells={anchor=west}}
        ]

        \addplot[red,mark=pentagon*,,mark size=2.5pt,thick,mark options={fill=white,draw=red,line width=1.0pt}] coordinates { (6,27.30) (12,28.11) (24, 28.96)
        };
        \addlegendentry{\scalebox{.9}{Baseline}}
        \addplot[kleinblue,mark=diamond*,mark size=2.5pt,thick,mark options={fill=white,draw=kleinblue,line width=1.0pt}] coordinates {(6,28.56) (12,29.35) (24, 29.78)};
        \addlegendentry{\scalebox{.9}{TranSFormer}}
        \end{axis}  
      }
     
  \end{tikzpicture}
  }
  \subfloat {
    \begin{tikzpicture}
      \scriptsize{
        \begin{axis}[
	 at={(0,0)},
      ymajorgrids,
      xmajorgrids,
      grid style=dashed,
      width=.26\textwidth,
      height=.225\textwidth,
      legend style={at={(0.31,0.15)}, anchor=south west},
      xlabel={\scriptsize{BPE operations (K)}},
      ylabel={\scriptsize{BLEU}},
      ylabel style={yshift=-2em},xlabel style={yshift=1.0em},
      yticklabel style={/pgf/number format/precision=1,/pgf/number format/fixed zerofill},
      ymin=25.8,ymax=29.2,
      ytick={26.0,27.0,28.0, 29.0},
      xmin=5,xmax=55,
      xtick={10, 30, 50},
      legend style={yshift=-6pt,xshift=-1em, legend plot pos=right,fill opacity=0.5, font={\tiny},cells={anchor=west}}
      ] 
      \addplot[red,mark=pentagon*,,mark size=2.5pt,thick,mark options={fill=white,draw=red,line width=1.0pt}] coordinates {(10,26.8) (32,27.40) (50,27.65)};
      \addlegendentry{\scalebox{.9}{{Baseline}}}

      \addplot[kleinblue,mark=diamond*,mark size=2.5pt,thick,mark options={fill=white,draw=kleinblue,line width=1.0pt}] coordinates {(10,28.22) (32,28.56) (50,28.74)
      };
      \addlegendentry{\scalebox{.9}{TranSFormer}}

      \end{axis}
       }
    \end{tikzpicture}
  }
    \caption{BLEU against differnt BPE merging operations and model depths.}
    \label{fig:analysis}
\end{figure}

\paragraph{Ablations on interactions in different levels}
Our default configuration permits the model to allocate interactions at each encoder layer. It is beneficial to determine how interaction frequency impacts the performance. Table \ref{tab:ablation_fusion_methods} compares various interaction frequencies at different levels, including exclusively at the final encoder block, the bottom three blocks, the top three blocks and every two blocks. The experiments were conducted on WMT En-De. It is evident that the default configuration delivers optimal performance. Interactions conducted at the top three blocks demonstrate superior results compared to those at the bottom three blocks. Furthermore, performing fusion solely at the last encoder block proves insufficient for the model to learn multiscale interactions effectively.

\paragraph{BLEU \textit{v.s.} Depth and BPE mergings}
Figure \ref{fig:analysis} plots the performance against model depths and BPE merging operations. The proposed TranSFormer architecture demonstrates consistent performance improvements as a result of increased encoder depth. Furthermore, an empirical evaluation of the TranSFormer against various byte-pair encoding (BPE) operations \cite{sennrich-subword-neural}, on the slow branch of the model yields a statistically significant average gain of $1.1$ BLEU scores over the Slow only baseline.
  \begin{figure}
    \centering
    \begin{tikzpicture}
      \scriptsize{
      \begin{axis}[
    ymajorgrids=true,
        xmajorgrids=true,
        grid style=dashdotted,
        ybar,
        enlarge x limits=0.2,
        height=.5\linewidth,
        width=0.9\linewidth,
        bar width=1.4em,
        symbolic x coords = {{Negation},{Past},{P-Gender},{P-Number}},
        ymin=90,
        ymax=100.5,
        ytick = {90.0,92.0,94.0,96.0,98.0,100.0},
        xtick = data,
        xticklabel style={/pgf/number format/fixed,/pgf/number format/fixed zerofill,/pgf/number format/precision=1},]
        \addplot[fill=blue!30, draw=blue!50] coordinates {({Negation},97.2) ({Past},90.8) ({P-Gender},99.2) ({P-Number},99.2)};
        \addplot[fill=teal!50, draw=teal!70, xshift=-0.15em] coordinates {({Negation},99.2) ({Past},93.3) ({P-Gender},100) ({P-Number},100)};

    \node [minimum size=0.1em, fill=blue!30, draw=blue!50] (n1) at (0.5cm,2cm){};
    \node [anchor=west,] (w1) at ([xshift=0.3em]n1.east){Baseline};
    \node [anchor=north,minimum size=0.1em, fill=teal!50, draw=teal!70] (n2) at ([yshift=-0.4em]n1.south){};
    \node [anchor=west,] (w2) at ([xshift=0.3em]n2.east){TranSFormer};

      \end{axis}
    }
    \end{tikzpicture}
    \caption{Results on MorphEval. Where P-Gender and P-Number stands for pron2nouns-gender and pron2nouns-number.}
    \label{fig:error-statistics}
    \end{figure}
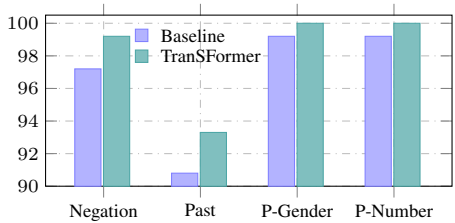

\begin{table}
    \centering
    \small
    \begin{tabular}{l|l|l}
         \multicolumn{1}{c|}{\#} & Model  & FLOPs  \\
        \Xhline{1.0pt}
        1 & Baseline          & 1.1G FLOPs \\
        2 & TranSFormer       & 1.4G FLOPs \\ 
        3 & Baseline (Big)    & 3.9G FLOPs \\
        4 & TranSFormer (Big) & 5.0G FLOPs \\
    \end{tabular}
    \caption{\label{tab:flops}
    Comparison of FLOPs of different models.}
\end{table}

\paragraph{Low resource setting and morphological evaluation}
\citet{li-etal-2021-char} has shown that character-level systems are better at handling morphological phenomena and show strong performance in low-resource scenarios than subword-level systems. Consequently, we evaluate how TranSFormer behaves at these scenarios.
For the low-resource setting, we randomly select subsets of 50K, 500K, and 1000K from the WMT En-De training corpus. TranSFormer achieves respective BLEU scores of 11.87, 22.75, and 25.30, while the character-only and subword-only Transformers yield approximate scores of 10.50/7.00, 20.50/22.00, and 22.50/23.50. This empirical evidence demonstrates that TranSFormer effectively amalgamates the benefits of both character-level and subword-level features. Moreover, Figure \ref{fig:error-statistics} plots the performance on MorphEval\cite{burlot-yvon-2017-evaluating} benchmark. TranSFormer behaves better than subword solely in terms of Negation, Past, P-Gender and P-Number metrics.

\paragraph{Comparisons in Efficiency}
Table \ref{tab:flops} compares the FLOPs between baseline and our TranSFormer both in base and big configurations. Due to the light computation cost of the fast branch, TranSFormer only brings additional 0.3G/1.1G FLOPS in base/big configurations, respectively. Note that the bulk of the additional computational cost is associated with the upsampling/downsampling operations within the cross-granularity attention mechanism. This process aligns the hidden size between the two representations.

\section{Conclusions}
In this work, we comprehensively leverage the potential of character-level features in multiscale sequence models while preserving high computational efficiency. To accomplish this, we propose a Slow-Fast Transformer architecture consisting of two branches in the encoder. The slow branch, akin to the vanilla Transformer, handles subword-level features, while the fast branch captures fine-grained correlations among characters. By leveraging the complementary features provided by the fast branch, our TranSFormer demonstrates consistent improvements in BLEU scores on three widely-used machine translation benchmarks. Further in-depth analyses demonstrate the effectiveness of the TranSFormer and its potential as a universal multiscale learning framework.

\section*{Acknowledgments}
This work was supported in part by the National Science Foundation of China (No. 62276056), the National Key R\&D Program of China, the China HTRD Center Project (No. 2020AAA0107904), the Natural Science Foundation of Liaoning Province of China (2022-KF-16-01), the Yunnan Provincial Major Science and Technology Special Plan Projects (No. 202103AA080015), the Fundamental Research Funds for the Central Universities (Nos. N2216016, N2216001, and N2216002), and the Program of Introducing Talents of Discipline to Universities, Plan 111 (No. B16009).

\section*{Limitations}
The proposed TranSFormer architecture employs a two-branch design, which separately encodes character-level and subword-level features. Our original design of the proposed cross-granularity attention is to acknowledge the correlation between subwords and characters that belong to the same word. For example, a cross-granularity Gaussian distribution to let subwords pay more attention to the corresponding characters. However, the variability of word boundary information across sentences presents a challenge in effectively batching them and achieving high computational efficiency. This is an area of ongoing research, and will be the focus of future work. On the other hand, our current evaluation of the TranSFormer architecture is limited to machine translation tasks. It is worth exploring the potential of TranSFormer in optimizing character sequences on natural language understanding tasks and other sequence generation tasks, such as abstractive summarization. These tasks are more challenging in terms of encoding longer sequences, but we believe that TranSFormer can serve as a versatile backbone. We aim to verify its effectiveness on these tasks in the future.

\bibliography{custom}

\begin{thebibliography}{36}
\expandafter\ifx\csname natexlab\endcsname\relax\def\natexlab#1{#1}\fi

\bibitem[{Ahmed et~al.(2017)Ahmed, Keskar, and Socher}]{Ahmed2017weighted}
Karim Ahmed, Nitish~Shirish Keskar, and Richard Socher. 2017.
\newblock \href {http://arxiv.org/abs/1711.02132} {Weighted transformer network
  for machine translation}.
\newblock \emph{CoRR}, abs/1711.02132.

\bibitem[{Burlot and Yvon(2017)}]{burlot-yvon-2017-evaluating}
Franck Burlot and Fran{\c{c}}ois Yvon. 2017.
\newblock \href {https://doi.org/10.18653/v1/W17-4705} {Evaluating the
  morphological competence of machine translation systems}.
\newblock In \emph{Proceedings of the Second Conference on Machine
  Translation}, pages 43--55, Copenhagen, Denmark. Association for
  Computational Linguistics.

\bibitem[{Cherry et~al.(2018)Cherry, Foster, Bapna, Firat, and
  Macherey}]{cherry-etal-2018-revisiting}
Colin Cherry, George Foster, Ankur Bapna, Orhan Firat, and Wolfgang Macherey.
  2018.
\newblock \href {https://doi.org/10.18653/v1/D18-1461} {Revisiting
  character-based neural machine translation with capacity and compression}.
\newblock In \emph{Proceedings of the 2018 Conference on Empirical Methods in
  Natural Language Processing}, pages 4295--4305, Brussels, Belgium.
  Association for Computational Linguistics.

\bibitem[{Fan et~al.(2020)Fan, Xie, Xia, Wu, Qin, Li, and
  Liu}]{Fan-multibranch-attentive}
Yang Fan, Shufang Xie, Yingce Xia, Lijun Wu, Tao Qin, Xiang{-}Yang Li, and
  Tie{-}Yan Liu. 2020.
\newblock \href {https://arxiv.org/abs/2006.10270} {Multi-branch attentive
  transformer}.
\newblock \emph{CoRR}, abs/2006.10270.

\bibitem[{Feichtenhofer et~al.(2019)Feichtenhofer, Fan, Malik, and
  He}]{Christoph2019slowfast}
Christoph Feichtenhofer, Haoqi Fan, Jitendra Malik, and Kaiming He. 2019.
\newblock Slowfast networks for video recognition.
\newblock In \emph{2019 {IEEE/CVF} International Conference on Computer Vision,
  {ICCV} 2019, Seoul, Korea (South), October 27 - November 2, 2019}, pages
  6201--6210. {IEEE}.

\bibitem[{Gao et~al.(2020)Gao, Nikolov, Hu, and
  Hahnloser}]{gao-etal-2020-character}
Yingqiang Gao, Nikola~I. Nikolov, Yuhuang Hu, and Richard~H.R. Hahnloser. 2020.
\newblock \href {https://doi.org/10.18653/v1/2020.acl-main.145}
  {Character-level translation with self-attention}.
\newblock In \emph{Proceedings of the 58th Annual Meeting of the Association
  for Computational Linguistics}, pages 1591--1604, Online. Association for
  Computational Linguistics.

\bibitem[{Gulati et~al.(2020)Gulati, Qin, Chiu, Parmar, Zhang, Yu, Han, Wang,
  Zhang, Wu, and Pang}]{gulati2020conformer}
Anmol Gulati, James Qin, Chung{-}Cheng Chiu, Niki Parmar, Yu~Zhang, Jiahui Yu,
  Wei Han, Shibo Wang, Zhengdong Zhang, Yonghui Wu, and Ruoming Pang. 2020.
\newblock \href {https://doi.org/10.21437/Interspeech.2020-3015} {Conformer:
  Convolution-augmented transformer for speech recognition}.
\newblock In \emph{Interspeech 2020, 21st Annual Conference of the
  International Speech Communication Association, Virtual Event, Shanghai,
  China, 25-29 October 2020}, pages 5036--5040. {ISCA}.

\bibitem[{Guo et~al.(2020)Guo, Qiu, Liu, Xue, and Zhang}]{guo-2020-msan}
Qipeng Guo, Xipeng Qiu, Pengfei Liu, Xiangyang Xue, and Zheng Zhang. 2020.
\newblock \href {https://aaai.org/ojs/index.php/AAAI/article/view/6290}
  {Multi-scale self-attention for text classification}.
\newblock In \emph{The Thirty-Fourth {AAAI} Conference on Artificial
  Intelligence, {AAAI} 2020, The Thirty-Second Innovative Applications of
  Artificial Intelligence Conference, {IAAI} 2020, The Tenth {AAAI} Symposium
  on Educational Advances in Artificial Intelligence, {EAAI} 2020, New York,
  NY, USA, February 7-12, 2020}, pages 7847--7854. {AAAI} Press.

\bibitem[{Han et~al.(2021)Han, Xiao, Wu, Guo, Xu, and
  Wang}]{han2021transformer}
Kai Han, An~Xiao, Enhua Wu, Jianyuan Guo, Chunjing Xu, and Yunhe Wang. 2021.
\newblock Transformer in transformer.
\newblock \emph{arXiv preprint arXiv:2103.00112}.

\bibitem[{Hao et~al.(2019)Hao, Wang, Shi, Zhang, and Tu}]{hao-etal-2019-multi}
Jie Hao, Xing Wang, Shuming Shi, Jinfeng Zhang, and Zhaopeng Tu. 2019.
\newblock \href {https://doi.org/10.18653/v1/D19-1082} {Multi-granularity
  self-attention for neural machine translation}.
\newblock In \emph{Proceedings of the 2019 Conference on Empirical Methods in
  Natural Language Processing and the 9th International Joint Conference on
  Natural Language Processing (EMNLP-IJCNLP)}, pages 887--897, Hong Kong,
  China. Association for Computational Linguistics.

\bibitem[{Hassan et~al.(2018)Hassan, Aue, Chen, Chowdhary, Clark, Federmann,
  Huang, Junczys-Dowmunt, Lewis, Li et~al.}]{hassan2018achieving}
Hany Hassan, Anthony Aue, Chang Chen, Vishal Chowdhary, Jonathan Clark,
  Christian Federmann, Xuedong Huang, Marcin Junczys-Dowmunt, William Lewis,
  Mu~Li, et~al. 2018.
\newblock Achieving human parity on automatic chinese to english news
  translation.
\newblock \emph{arXiv preprint arXiv:1803.05567}.

\bibitem[{Kasai et~al.(2020)Kasai, Cross, Ghazvininejad, and
  Gu}]{kasai2020disco}
Jungo Kasai, James Cross, Marjan Ghazvininejad, and Jiatao Gu. 2020.
\newblock \href {http://proceedings.mlr.press/v119/kasai20a.html}
  {Non-autoregressive machine translation with disentangled context
  transformer}.
\newblock In \emph{Proceedings of the 37th International Conference on Machine
  Learning, {ICML} 2020, 13-18 July 2020, Virtual Event}, volume 119 of
  \emph{Proceedings of Machine Learning Research}, pages 5144--5155. {PMLR}.

\bibitem[{Kingma and Ba(2015)}]{kingma2014adam}
Diederik~P. Kingma and Jimmy Ba. 2015.
\newblock \href {http://arxiv.org/abs/1412.6980} {Adam: {A} method for
  stochastic optimization}.
\newblock In \emph{3rd International Conference on Learning Representations,
  {ICLR} 2015, San Diego, CA, USA, May 7-9, 2015, Conference Track
  Proceedings}.

\bibitem[{Kipf and Welling(2017)}]{Kipf2017gcn}
Thomas~N. Kipf and Max Welling. 2017.
\newblock \href {https://openreview.net/forum?id=SJU4ayYgl} {Semi-supervised
  classification with graph convolutional networks}.
\newblock In \emph{5th International Conference on Learning Representations,
  {ICLR} 2017, Toulon, France, April 24-26, 2017, Conference Track
  Proceedings}. OpenReview.net.

\bibitem[{Kudo and Richardson(2018)}]{kudo-richardson-2018-sentencepiece}
Taku Kudo and John Richardson. 2018.
\newblock \href {https://doi.org/10.18653/v1/D18-2012} {{S}entence{P}iece: A
  simple and language independent subword tokenizer and detokenizer for neural
  text processing}.
\newblock In \emph{Proceedings of the 2018 Conference on Empirical Methods in
  Natural Language Processing: System Demonstrations}, pages 66--71, Brussels,
  Belgium. Association for Computational Linguistics.

\bibitem[{Lee et~al.(2017)Lee, Cho, and Hofmann}]{lee-etal-2017-fully}
Jason Lee, Kyunghyun Cho, and Thomas Hofmann. 2017.
\newblock \href {https://doi.org/10.1162/tacl\_a\_00067} {Fully character-level
  neural machine translation without explicit segmentation}.
\newblock \emph{Transactions of the Association for Computational Linguistics},
  5:365--378.

\bibitem[{Lee et~al.(2018)Lee, Mansimov, and Cho}]{lee-etal-2018-deterministic}
Jason Lee, Elman Mansimov, and Kyunghyun Cho. 2018.
\newblock \href {https://doi.org/10.18653/v1/D18-1149} {Deterministic
  non-autoregressive neural sequence modeling by iterative refinement}.
\newblock In \emph{Proceedings of the 2018 Conference on Empirical Methods in
  Natural Language Processing}, pages 1173--1182, Brussels, Belgium.
  Association for Computational Linguistics.

\bibitem[{Li et~al.(2022{\natexlab{a}})Li, Du, Zhou, Jing, Zhou, Zeng, Xiao,
  Zhu, Liu, and Zhang}]{li-etal-2022-ode}
Bei Li, Quan Du, Tao Zhou, Yi~Jing, Shuhan Zhou, Xin Zeng, Tong Xiao, JingBo
  Zhu, Xuebo Liu, and Min Zhang. 2022{\natexlab{a}}.
\newblock \href {https://doi.org/10.18653/v1/2022.acl-long.571} {{ODE}
  transformer: An ordinary differential equation-inspired model for sequence
  generation}.
\newblock In \emph{Proceedings of the 60th Annual Meeting of the Association
  for Computational Linguistics (Volume 1: Long Papers)}, pages 8335--8351,
  Dublin, Ireland. Association for Computational Linguistics.

\bibitem[{Li et~al.(2022{\natexlab{b}})Li, Zheng, Jing, Jiao, Xiao, and
  Zhu}]{li2022multiscale}
Bei Li, Tong Zheng, Yi~Jing, Chengbo Jiao, Tong Xiao, and Jingbo Zhu.
  2022{\natexlab{b}}.
\newblock Learning multiscale transformer models for sequence generation.
\newblock In \emph{International Conference on Machine Learning, {ICML} 2022,
  17-23 July 2022, Baltimore, Maryland, {USA}}, pages 13225--13241. {PMLR}.

\bibitem[{Li et~al.(2021)Li, Shen, Huang, Dai, and Chen}]{li-etal-2021-char}
Jiahuan Li, Yutong Shen, Shujian Huang, Xinyu Dai, and Jiajun Chen. 2021.
\newblock \href {https://doi.org/10.18653/v1/2021.acl-short.69} {When is char
  better than subword: A systematic study of segmentation algorithms for neural
  machine translation}.
\newblock In \emph{Proceedings of the 59th Annual Meeting of the Association
  for Computational Linguistics and the 11th International Joint Conference on
  Natural Language Processing (Volume 2: Short Papers)}, pages 543--549,
  Online. Association for Computational Linguistics.

\bibitem[{Liu et~al.(2020)Liu, Gu, Goyal, Li, Edunov, Ghazvininejad, Lewis, and
  Zettlemoyer}]{liu-etal-2020-multilingual-denoising}
Yinhan Liu, Jiatao Gu, Naman Goyal, Xian Li, Sergey Edunov, Marjan
  Ghazvininejad, Mike Lewis, and Luke Zettlemoyer. 2020.
\newblock \href {https://doi.org/10.1162/tacl\_a\_00343} {Multilingual
  denoising pre-training for neural machine translation}.
\newblock \emph{Transactions of the Association for Computational Linguistics},
  8:726--742.

\bibitem[{Mehta et~al.(2020)Mehta, Ghazvininejad, Iyer, Zettlemoyer, and
  Hajishirzi}]{mehta2020delight}
Sachin Mehta, Marjan Ghazvininejad, Srinivasan Iyer, Luke Zettlemoyer, and
  Hannaneh Hajishirzi. 2020.
\newblock \href {https://arxiv.org/abs/2008.00623} {Delight: Very deep and
  light-weight transformer}.
\newblock \emph{ArXiv preprint}, abs/2008.00623.

\bibitem[{Ott et~al.(2019)Ott, Edunov, Baevski, Fan, Gross, Ng, Grangier, and
  Auli}]{ott2019fairseq}
Myle Ott, Sergey Edunov, Alexei Baevski, Angela Fan, Sam Gross, Nathan Ng,
  David Grangier, and Michael Auli. 2019.
\newblock \href {https://doi.org/10.18653/v1/N19-4009} {fairseq: A fast,
  extensible toolkit for sequence modeling}.
\newblock In \emph{Proceedings of the 2019 Conference of the North {A}merican
  Chapter of the Association for Computational Linguistics (Demonstrations)},
  pages 48--53, Minneapolis, Minnesota. Association for Computational
  Linguistics.

\bibitem[{Ott et~al.(2018)Ott, Edunov, Grangier, and Auli}]{ott2018scaling}
Myle Ott, Sergey Edunov, David Grangier, and Michael Auli. 2018.
\newblock \href {https://doi.org/10.18653/v1/W18-6301} {Scaling neural machine
  translation}.
\newblock In \emph{Proceedings of the Third Conference on Machine Translation:
  Research Papers}, pages 1--9, Brussels, Belgium. Association for
  Computational Linguistics.

\bibitem[{Sennrich et~al.(2016)Sennrich, Haddow, and
  Birch}]{sennrich-subword-neural}
Rico Sennrich, Barry Haddow, and Alexandra Birch. 2016.
\newblock \href {https://doi.org/10.18653/v1/P16-1162} {Neural machine
  translation of rare words with subword units}.
\newblock In \emph{Proceedings of the 54th Annual Meeting of the Association
  for Computational Linguistics (Volume 1: Long Papers)}, pages 1715--1725,
  Berlin, Germany. Association for Computational Linguistics.

\bibitem[{Shaw et~al.(2018)Shaw, Uszkoreit, and Vaswani}]{shaw-etal-2018-self}
Peter Shaw, Jakob Uszkoreit, and Ashish Vaswani. 2018.
\newblock \href {https://doi.org/10.18653/v1/N18-2074} {Self-attention with
  relative position representations}.
\newblock In \emph{Proceedings of the 2018 Conference of the North {A}merican
  Chapter of the Association for Computational Linguistics: Human Language
  Technologies, Volume 2 (Short Papers)}, pages 464--468, New Orleans,
  Louisiana. Association for Computational Linguistics.

\bibitem[{Vaswani et~al.(2017)Vaswani, Shazeer, Parmar, Uszkoreit, Jones,
  Gomez, Kaiser, and Polosukhin}]{vaswani2017attention}
Ashish Vaswani, Noam Shazeer, Niki Parmar, Jakob Uszkoreit, Llion Jones,
  Aidan~N. Gomez, Lukasz Kaiser, and Illia Polosukhin. 2017.
\newblock \href
  {https://proceedings.neurips.cc/paper/2017/hash/3f5ee243547dee91fbd053c1c4a845aa-Abstract.html}
  {Attention is all you need}.
\newblock In \emph{Advances in Neural Information Processing Systems 30: Annual
  Conference on Neural Information Processing Systems 2017, December 4-9, 2017,
  Long Beach, CA, {USA}}, pages 5998--6008.

\bibitem[{Voita et~al.(2018)Voita, Serdyukov, Sennrich, and
  Titov}]{voita-etal-2018-context}
Elena Voita, Pavel Serdyukov, Rico Sennrich, and Ivan Titov. 2018.
\newblock \href {https://doi.org/10.18653/v1/P18-1117} {Context-aware neural
  machine translation learns anaphora resolution}.
\newblock In \emph{Proceedings of the 56th Annual Meeting of the Association
  for Computational Linguistics (Volume 1: Long Papers)}, pages 1264--1274,
  Melbourne, Australia. Association for Computational Linguistics.

\bibitem[{Wang et~al.(2020)Wang, Li, Khabsa, Fang, and Ma}]{wang2020linformer}
Sinong Wang, Belinda~Z Li, Madian Khabsa, Han Fang, and Hao Ma. 2020.
\newblock Linformer: Self-attention with linear complexity.
\newblock \emph{arXiv preprint arXiv:2006.04768}.

\bibitem[{Wu et~al.(2020)Wu, Xie, Xia, Fan, Lai, Qin, and Liu}]{wu2020mixed}
Lijun Wu, Shufang Xie, Yingce Xia, Yang Fan, Jian{-}Huang Lai, Tao Qin, and
  Tie{-}Yan Liu. 2020.
\newblock \href {http://proceedings.mlr.press/v119/wu20e.html} {Sequence
  generation with mixed representations}.
\newblock In \emph{Proceedings of the 37th International Conference on Machine
  Learning, {ICML} 2020, 13-18 July 2020, Virtual Event}, volume 119 of
  \emph{Proceedings of Machine Learning Research}, pages 10388--10398. {PMLR}.

\bibitem[{Wu et~al.(2018)Wu, Wang, Liu, and Ma}]{wu-etal-2018-phrase}
Wei Wu, Houfeng Wang, Tianyu Liu, and Shuming Ma. 2018.
\newblock \href {https://doi.org/10.18653/v1/D18-1408} {Phrase-level
  self-attention networks for universal sentence encoding}.
\newblock In \emph{Proceedings of the 2018 Conference on Empirical Methods in
  Natural Language Processing}, pages 3729--3738, Brussels, Belgium.
  Association for Computational Linguistics.

\bibitem[{Xue et~al.(2022)Xue, Barua, Constant, Al-Rfou, Narang, Kale, Roberts,
  and Raffel}]{xue-etal-2022-byt5}
Linting Xue, Aditya Barua, Noah Constant, Rami Al-Rfou, Sharan Narang, Mihir
  Kale, Adam Roberts, and Colin Raffel. 2022.
\newblock \href {https://doi.org/10.1162/tacl\_a\_00461} {{B}y{T}5: Towards a
  token-free future with pre-trained byte-to-byte models}.
\newblock \emph{Transactions of the Association for Computational Linguistics},
  10:291--306.

\bibitem[{Yan et~al.(2020)Yan, Meng, and Zhou}]{yan-etal-2020-multi}
Jianhao Yan, Fandong Meng, and Jie Zhou. 2020.
\newblock \href {https://doi.org/10.18653/v1/2020.emnlp-main.77} {Multi-unit
  transformers for neural machine translation}.
\newblock In \emph{Proceedings of the 2020 Conference on Empirical Methods in
  Natural Language Processing (EMNLP)}, pages 1047--1059, Online. Association
  for Computational Linguistics.

\bibitem[{Yang et~al.(2018)Yang, Tu, Wong, Meng, Chao, and
  Zhang}]{yang-etal-2018-modeling}
Baosong Yang, Zhaopeng Tu, Derek~F. Wong, Fandong Meng, Lidia~S. Chao, and Tong
  Zhang. 2018.
\newblock \href {https://doi.org/10.18653/v1/D18-1475} {Modeling localness for
  self-attention networks}.
\newblock In \emph{Proceedings of the 2018 Conference on Empirical Methods in
  Natural Language Processing}, pages 4449--4458, Brussels, Belgium.
  Association for Computational Linguistics.

\bibitem[{Yang et~al.(2019)Yang, Wang, Wong, Chao, and
  Tu}]{yang-etal-2019-convolutional}
Baosong Yang, Longyue Wang, Derek~F. Wong, Lidia~S. Chao, and Zhaopeng Tu.
  2019.
\newblock \href {https://doi.org/10.18653/v1/N19-1407} {Convolutional
  self-attention networks}.
\newblock In \emph{Proceedings of the 2019 Conference of the North {A}merican
  Chapter of the Association for Computational Linguistics: Human Language
  Technologies, Volume 1 (Long and Short Papers)}, pages 4040--4045,
  Minneapolis, Minnesota. Association for Computational Linguistics.

\bibitem[{Zhao et~al.(2019)Zhao, Sun, Xu, Zhang, and Luo}]{Zhao-2019-MUSE}
Guangxiang Zhao, Xu~Sun, Jingjing Xu, Zhiyuan Zhang, and Liangchen Luo. 2019.
\newblock \href {http://arxiv.org/abs/1911.09483} {{MUSE:} parallel multi-scale
  attention for sequence to sequence learning}.
\newblock \emph{CoRR}, abs/1911.09483.

\end{thebibliography}
\bibliographystyle{acl_natbib}

\end{document}